%% file: main.tex
\documentclass[sigconf]{acmart}
\AtBeginDocument{%
  }

\setcopyright{acmlicensed}
\copyrightyear{2026}
\acmYear{2026}
\setcopyright{cc}
\setcctype{by}
\acmConference[MM '26]{Proceedings of the 34th ACM International Conference on Multimedia}{November 10--14, 2026}{Rio de Janeiro, Brazil}
\acmBooktitle{Proceedings of the 34th ACM International Conference on Multimedia (MM '26), November 10--14, 2026, Rio de Janeiro, Brazil}
\acmDOI{10.1145/3767308.3835649}
\acmISBN{979-8-4007-2213-4/2026/11}

\usepackage{subcaption}
\usepackage[table]{xcolor}
\usepackage{algorithm}
\usepackage{algpseudocode}
\begin{document}

\title{BRACE: Taming Sharp Irregularities via Barycentric Rational Forecasting for Fast Diffusion Transformers Inference}

\author{Jinlong Yang}
\authornote{Jinlong Yang and Jinke Wu contributed equally to this research.}
\email{2023141461083@stu.scu.edu.cn}
\orcid{0009-0000-1809-850X}
\affiliation{%
  \institution{Sichuan University}
  \city{Chengdu}
  \country{China}
}

\author{Jinke Wu}
\email{w\_rmsl@stu.scu.edu.cn}
\authornotemark[1]
\affiliation{%
  \institution{Sichuan University}
  \city{Chengdu}
  \country{China}
}

\author{Lizilin}
\email{2023141461067@stu.scu.edu.cn}
\affiliation{%
  \institution{Sichuan University}
  \city{Chengdu}
  \country{China}
}

\author{Yao Zhou}
\correspondingauthor
\authornote{Corresponding author.}
\email{yaozhou@scu.edu.cn}
\affiliation{%
  \institution{School of Artificial Intelligence, Sichuan University}
  \city{Chengdu}
  \country{China}
}

\renewcommand{\shortauthors}{Yang et al.}

\begin{abstract}
Diffusion Transformers (DiTs) have demonstrated exceptional performance in high-fidelity image and video generation. To alleviate their massive computational overhead, temporal feature caching has been proposed to bypass redundant computations. However, existing cache-then-forecast methods driven by derivative-based polynomials often cause severe quality degradation under high acceleration due to unstable long-step predictions. To address this bottleneck, we propose Barycentric Rational Forecasting with Chebyshev Enhancement (BRACE). Motivated by the observation that DiT feature trajectories are globally smooth yet frequently exhibit sharp irregularities and local non-smoothness, BRACE shifts the paradigm from derivative-driven polynomial extrapolation to feature-driven rational forecasting. Specifically, it maintains a local sliding window to cache sparse historical features and leverages adapted Chebyshev weights to formulate a barycentric rational function, directly aggregating these raw features to ensure numerical stability. Extensive experiments demonstrate that BRACE achieves state-of-the-art quality–efficiency trade-offs across various DiT architectures with negligible computational overhead. Further details are available on our
\href{https://youngkinlon.github.io/BRACE-Taming-Sharp-Irregularities-via-Barycentric-Rational-Forecasting-for-Fast-DiT-Inference/}{project page}.
\end{abstract}



\maketitle
\input{sec_1_intro}
\input{sec_2_related_work}
\input{sec_3_Method}
\input{sec_4_Experiments}
\input{sec_5_conclusion}
\begin{acks}
This work was supported by the National Natural Science Foundation of China under Grant No.~62376172.
\end{acks}
\bibliographystyle{ACM-Reference-Format}
\bibliography{main}   
\end{document}

%% file: sec_1_intro.tex
\section{Introduction}
Diffusion Models (DMs)~\cite{ho2020denoisingdiffusionprobabilisticmodels} have become the dominant generative modeling paradigm across diverse modalities including images~\cite{chen2023pixartalphafasttrainingdiffusion,cai2025hidreami1highefficientimagegenerative,flux2024}, videos~\cite{kong2024hunyuanvideo,wan2025}, and audio~\cite{kong2021diffwaveversatilediffusionmodel,liu2023audioldm}, driven by their exceptional high-fidelity content synthesis capability. The recent architectural shift to Diffusion Transformers (DiTs)~\cite{peebles2023scalablediffusionmodelstransformers} has further amplified this synthesis performance, coming at the cost of prohibitive computational overhead. Specifically, the iterative sampling process requires sequential forward passes through large-scale models, resulting in high latency that severely limits real-time interactive deployment and high-throughput generation.

To alleviate this bottleneck, feature caching has emerged as a prominent training-free acceleration strategy. Built on the strong temporal consistency of intermediate hidden representations, traditional caching approaches directly reuse features from preceding timesteps, but this 'cache-and-reuse' paradigm inherently degrades generation quality under long skip intervals as feature similarity diminishes with extended temporal distances~\cite{selvaraju2024FORAfastforwardcachingdiffusion,ma2023deepcacheacceleratingdiffusionmodels}. Recently, this research direction has evolved from static feature reuse to proactive forecasting: representative frameworks such as TaylorSeer~\cite{liu2025reusingforecastingacceleratingdiffusion} and HiCache~\cite{feng2026hicachepluginscaledhermiteupgrade} construct derivative-driven polynomial paradigms via Taylor expansions and dual-scale Hermite-inspired extrapolation, respectively. These methods achieve promising acceleration results by approximating derivatives with finite differences of sparse historical features for future state prediction.

\begin{figure}[t] 
    \centering
    \begin{subfigure}[t]{0.48\linewidth}  
        \centering
        \includegraphics[width=\linewidth]{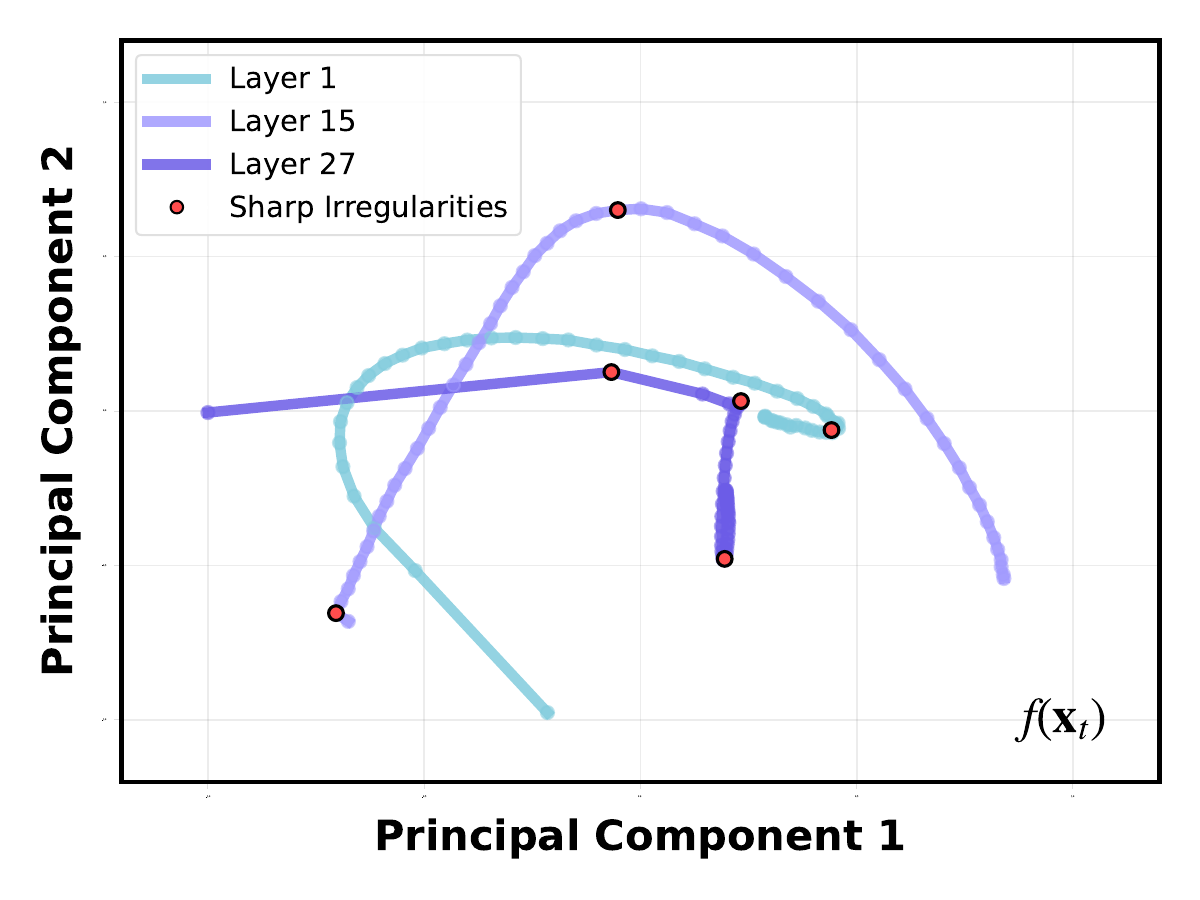}
        \caption{Manifold Irregularities} 
        \label{fig:trajectory}
    \end{subfigure}
    \hfill 
    \begin{subfigure}[t]{0.48\linewidth}
        \centering
        \includegraphics[width=\linewidth]{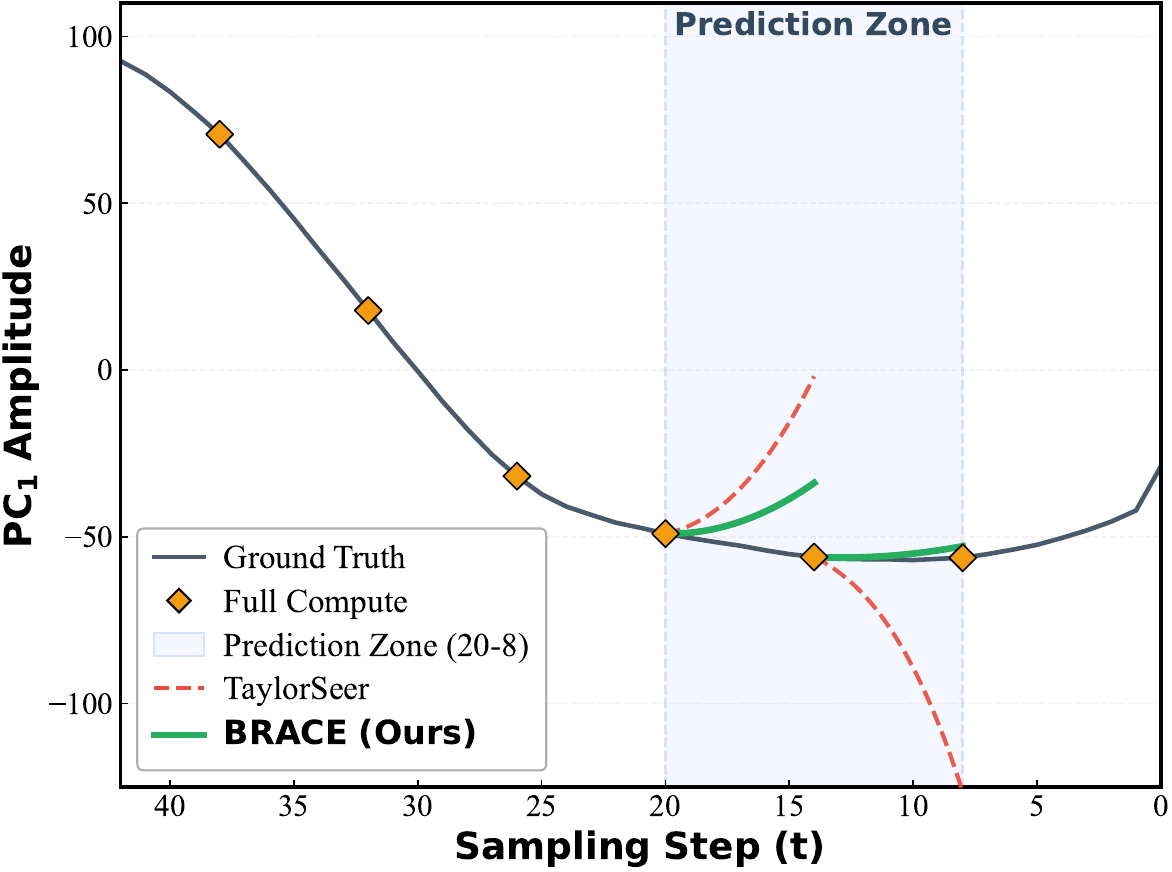}
        \caption{Prediction Drift Analysis} 
        \label{fig:drift_analysis}
    \end{subfigure}
    \caption{Feature geometric complexity. 
    (a) Trajectories exhibit sharp irregularities across all layers. 
    (b) At Layer 15, TaylorSeer diverges at inflection points, while BRACE robustly tracks the actual manifold.}
    \label{fig:geometric_analysis}
    \Description{Two side-by-side plots analyzing feature geometry. Left plot shows PCA trajectories of features with sharp, irregular changes across layers. Right plot compares predicted feature trajectories at layer 15: TaylorSeer deviates significantly at inflection points, while BRACE closely follows the ground truth manifold.}
\end{figure}

While innovative, this derivative-driven polynomial paradigm fundamentally relies on the accuracy of instantaneous derivatives estimated from discrete historical data, and suffers from the inherent instability of polynomial extrapolation over large skip intervals. To clarify the core challenges of DiT acceleration, we empirically analyze DiT feature evolution trajectories, and reveal an inherent geometric property of DiTs: their feature trajectories, while globally smooth, frequently exhibit sharp irregularities and local non-smoothness. This dual geometric nature is visualized in Figure~\ref{fig:geometric_analysis}(a), and aligns with phase transition phenomena in diffusion generative processes~\cite{lobashev2025hessian}, where the latent manifold undergoes rapid structural shifts across specific denoising intervals.

Numerically, these rapid transitions mimic stiff system behavior, with feature states changing drastically within extremely short intervals. This localized high curvature exposes two fundamental limitations of existing paradigms: first, finite-difference gradient estimates become highly unreliable at sharp inflection points; second, rigid polynomials lack the structural adaptability to model abrupt shifts, leading to severe extrapolation divergence outside the local window~\cite{floater2007barycentric}. When extrapolating across these high-frequency transition regions, derivative-centric polynomials inevitably overshoot or oscillate, accumulating severe truncation errors. As verified in Figure~\ref{fig:geometric_analysis}(b), derivative-driven polynomial methods like TaylorSeer~\cite{liu2025reusingforecastingacceleratingdiffusion} fail to capture these abrupt transitions due to the above deficiencies, resulting in severe prediction drift and loss of structural integrity.

In classical numerical analysis, it is well-established that rational extrapolation consistently outperforms its polynomial counterparts in modeling rapid transitions~\cite{STOER1966}, and barycentric rational interpolation uniquely enables explicit detection and elimination of unattainable points and spurious interior poles~\cite{berrut2004barycentric}. This stabilizing property in interpolation heuristically inspires us to extend this approach to causal prediction, enabling the accurate capture of complex DiT feature dynamics and the effective taming of trajectory irregularities. Driven by this motivation, BRACE (Barycentric Rational Forecasting with Chebyshev Enhancement) is introduced as a novel forecasting paradigm for structurally robust diffusion acceleration and superior generation quality.

BRACE shifts the forecasting paradigm from derivative-driven polynomial extrapolation to feature-driven rational forecasting. Specifically, rather than estimating unstable local derivatives, it directly aggregates raw historical features cached in a sparse local sliding window. To proactively suppress long-step oscillations, adapted Chebyshev weights are introduced and seamlessly unified within the barycentric rational framework to enable stable causal extrapolation. This highly expressive rational predictor replaces rigid polynomials to smoothly accommodate high-curvature dynamics, bridging the gap between the global smoothness and local geometric complexity of DiT feature trajectories. This structural robustness is critical for long-step acceleration: as step size increases, derivative-driven polynomials oscillate or diverge, while BRACE maintains numerical stability and high fidelity even under aggressive skip intervals, successfully capturing localized sharp transitions without overshooting, as shown in Figure~\ref{fig:geometric_analysis}(b).

In summary, our core contributions are as follows:
\begin{itemize}
\item \textbf{The BRACE Framework:} We propose BRACE, a training-free rational forecasting framework for efficient DiT inference. By synergizing a sparse historical cache with adapted Chebyshev weights within a barycentric rational framework, BRACE ensures superior numerical stability and tames abrupt transitions where traditional derivative-driven methods fail.
\item \textbf{Geometric Analysis of DiT Dynamics:} We reveal the dual geometric nature of DiT evolution: globally smooth trajectories frequently punctuated by sharp irregularities, and empirically validate that rational bases inherently outperform polynomials in fitting DiT feature trajectories, providing structural motivation for our approach.
\item \textbf{State-of-the-Art Acceleration:} Extensive evaluations across diverse architectures and modalities—including DiT-XL/2, FLUX.1-dev, and HunyuanVideo—demonstrate that BRACE consistently outperforms existing baselines, maintaining exceptional generation fidelity even under aggressive skip intervals.
\end{itemize}

%% file: sec_2_related_work.tex
\section{Related Work}

\subsection{Computation Reduction Strategies}
Previous efforts to accelerate diffusion models primarily focus on either reducing the number of sampling steps or compressing the denoising network itself. Step-reduction methods accelerate the generative process through efficient ODE solvers like DDIM~\cite{song2022denoisingdiffusionimplicitmodels} and DPM-Solver~\cite{lu2022dpmsolverfastodesolver,zhao2023unipcunifiedpredictorcorrectorframework,Lu_2025}, knowledge distillation~\cite{salimans2022progressivedistillationfastsampling,luo2023latentconsistencymodelssynthesizing,sauer2023adversarialdiffusiondistillation}, or few-step paradigms such as Rectified Flow~\cite{liu2022flowstraightfastlearning} and Consistency Models~\cite{song2023consistencymodels}. However, these approaches often degrade generation quality or require expensive retraining~\cite{ma2024sitexploringflowdiffusionbased}. Alternatively, network compression techniques like pruning~\cite{fang2023structuralpruningdiffusionmodels,yuan2024ditfastattn}, quantization~\cite{li2023qdiffusionquantizingdiffusionmodels,shang2023ptqdm}, and token merging~\cite{zhang2024tokenpruningcachingbetter,bolya2023tokenmergingfaststable} reduce FLOPs per step but entail a strict efficiency--expressivity trade-off, risking generation fidelity under aggressive compression. Consequently, developing a plug-and-play acceleration paradigm that dynamically exploits redundancy without compromising the original synthesis quality remains a critical open challenge across diverse model architectures.

\subsection{Feature Caching}
As a prominent training-free alternative, feature caching skips redundant intermediate computations. Existing methods generally fall into two paradigms: cache-then-reuse and cache-then-forecast.Reuse-based approaches typically choose to directly copy cached representations, spanning various spatial granularities from global spatial-temporal feature reuse (e.g., DeepCache~\cite{ma2023deepcacheacceleratingdiffusionmodels}, FORA~\cite{selvaraju2024FORAfastforwardcachingdiffusion}, PAB~\cite{zhao2024real}) to fine-grained token-level caching (e.g., ToCa~\cite{zou2025acceleratingdiffusiontransformerstokenwise}, Tokencache~\cite{lou2024token}, DaTo~\cite{zhang2024tokenpruningcachingbetter}). However, regardless of the spatial granularity, the temporal representation mismatch inevitably grows with larger skip intervals, fundamentally degrading generation quality.

Forecasting-based approaches, pioneered by TaylorSeer~\cite{liu2025reusingforecastingacceleratingdiffusion}, actively predict feature evolution to bridge this temporal gap. Recent advancements seek to refine this paradigm: HiCache~\cite{feng2026hicachepluginscaledhermiteupgrade} improves the standard Taylor expansion by introducing scaled Hermite coefficients, while FOCA~\cite{zheng2025forecastcalibratefeaturecaching} and SpeCA~\cite{liu2025speca} integrate verify-and-calibrate mechanisms into the forecasting framework to mitigate extrapolation drift. Nevertheless, because their underlying prediction modules inherently rely on volatile derivative estimates and rigid polynomial bases, they remain brittle near sharp transitions. Our method instead performs highly expressive barycentric rational extrapolation directly from raw historical features, ensuring robust trajectory alignment even under aggressive skip intervals.

%% file: sec_3_Method.tex
\section{Method}

\subsection{Preliminaries}
Diffusion models generate data by iteratively reversing a predefined forward noising process via a neural network $\epsilon_\theta(\mathbf{x}_t, t)$. Starting from Gaussian noise, the latent variable $\mathbf{x}_t$ is progressively refined over $T$ timesteps:
\begin{equation}
    \mathbf{x}_{t-1} = \frac{1}{\sqrt{\alpha_t}} \left( \mathbf{x}_t - \frac{1 - \alpha_t}{\sqrt{1 - \bar{\alpha}_t}} \epsilon_\theta(\mathbf{x}_t, t) \right) + \sigma_t \boldsymbol{\epsilon}.
    \label{eq:reverse_process}
\end{equation}
Requiring $T$ sequential model evaluations, the sampling process incurs massive computational overhead.

To mitigate this, \textit{forecast-based caching} approximates future intermediate feature maps $\mathbf{F}^l$ at layer $l$. Specifically, using a recursive discrete difference operator $\Delta$, the $m$-th order Taylor prediction for a skip interval $s$ is expressed as:
\begin{equation}
    \mathbf{F}_{\text{pred}}(\mathbf{x}^l_{t-s}) = \mathbf{F}(\mathbf{x}^l_t) + \sum_{i=1}^{m} \frac{\Delta^i \mathbf{F}(\mathbf{x}^l_t)}{i!} (-s)^i.
    \label{eq:taylor}
\end{equation}
According to Taylor's theorem~\cite{trefethenbau1997numerical}, the accuracy of this polynomial extrapolation is governed by the Lagrange remainder:
\begin{equation}
    \mathcal{R}_m = \frac{\mathbf{F}^{(m+1)}(\xi)}{(m+1)!} (-s)^{m+1}, \quad \xi \in [t-s, t].
    \label{eq:lagrange_remainder}
\end{equation}
As shown in Eq.~\ref{eq:lagrange_remainder}, the Taylor forecasting error $\mathcal{R}_m$ is theoretically bounded by the high-order derivative $\|\mathbf{F}^{(m+1)}(\xi)\|$ and the skip interval $s^{m+1}$. This formulation exposes a critical vulnerability: if intermediate features exhibit non-smooth dynamics, the resulting spikes in true derivatives will be severely amplified by large skip steps, risking catastrophic extrapolation divergence.

\subsection{Feature Trajectory Insight}
\begin{figure}[t]
    \centering
    \begin{subfigure}[b]{0.49\linewidth}
        \centering
        \includegraphics[width=\linewidth]{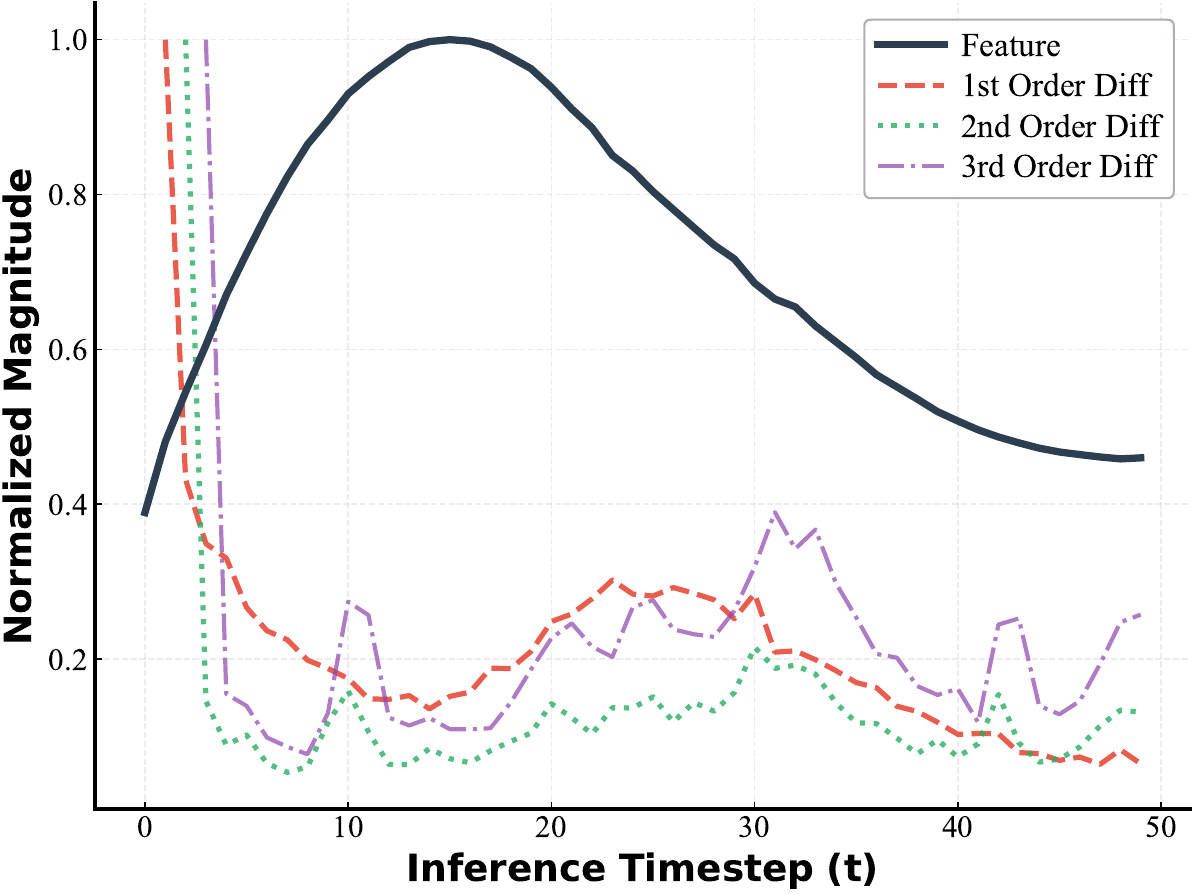}
        \caption{Numerical Instability of Differences}
        \label{fig:diff_instability}
    \end{subfigure}
    \hfill
    \begin{subfigure}[b]{0.49\linewidth}
        \centering
        \includegraphics[width=\linewidth]{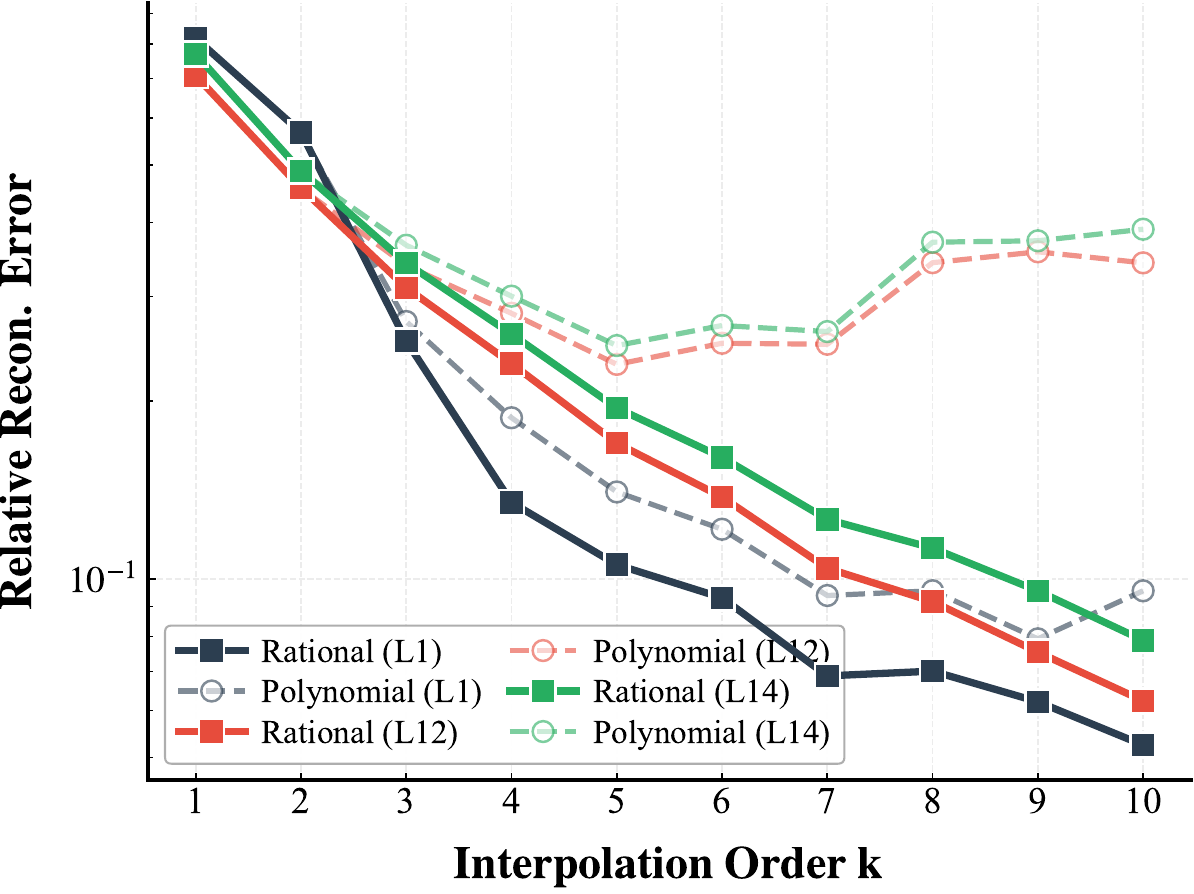}
        \caption{Rational vs. Polynomial Expressiveness}
        \label{fig:basis_comparison}
    \end{subfigure}
    \caption{DiT feature dynamics and basis expressiveness. 
    (a) Magnitude of finite differences. Localized non-smoothness leads to an explosion in $\|\mathbf{F}^{(m+1)}\|$ as the order increases. 
    (b) Interpolation error comparison. Unlike drifting polynomials, rational bases accommodate local singularities under identical historical constraints.}
    \label{fig:error_analysis}
    \Description{Two side-by-side analysis plots. Left plot shows the magnitude of finite differences in DiT features, demonstrating numerical instability from non-smooth changes. Right plot compares interpolation errors: rational basis methods maintain stability while polynomial methods show large prediction drifts.}
\end{figure}

Coupling this theoretical vulnerability with the actual dual nature of feature trajectories—globally smooth yet punctuated by sharp transitions—it becomes imperative to identify an optimal forecasting tool. To determine the most robust mathematical framework for such dynamics, we conduct two parallel empirical analyses focusing on the numerical stability of finite differences and intrinsic basis expressiveness, as depicted in Fig.~\ref{fig:error_analysis}.

As shown in Fig.~\ref{fig:error_analysis}(a), although raw feature trajectories appear globally smooth, their high-order finite differences produce erratic spikes at localized transitions. This empirically confirms the vulnerability identified in Eq.~\eqref{eq:lagrange_remainder}: localized non-smoothness triggers an explosion in $\|\mathbf{F}^{(m+1)}\|$. Consequently, traditional derivative-driven extrapolation relying on these unstable estimates suffers from severe error amplification over long steps.

Beyond numerical stability, rational bases theoretically surpass polynomials in complex function approximation~\cite{STOER1966}. Fig.~\ref{fig:error_analysis}(b) empirically validates their inherent optimality for feature dynamics: as the interpolation order increases, the polynomial baseline reveals a fundamental geometric mismatch, severely drifting at sharp trajectory turns instead of fitting them.

These empirical insights thus motivate a paradigm shift from derivative-driven polynomial estimation to feature-driven rational extrapolation. Rather than relying on high-order derivatives, our approach forecasts future states by directly aggregating stable historical features. Crucially, by replacing rigid polynomials with rational bases, the self-normalizing denominator inherently ``absorbs'' localized non-smoothness, thereby effectively preventing the error explosion characteristic of polynomial drift. Ultimately, this combination of derivative-free aggregation and rational structural robustness establishes a fundamentally more stable foundation for long-step DiT acceleration, ensuring high generative fidelity even across sharp feature transitions.

\subsection{The BRACE Framework}

\begin{figure*}[t]
\centering
\includegraphics[width=0.9\linewidth]{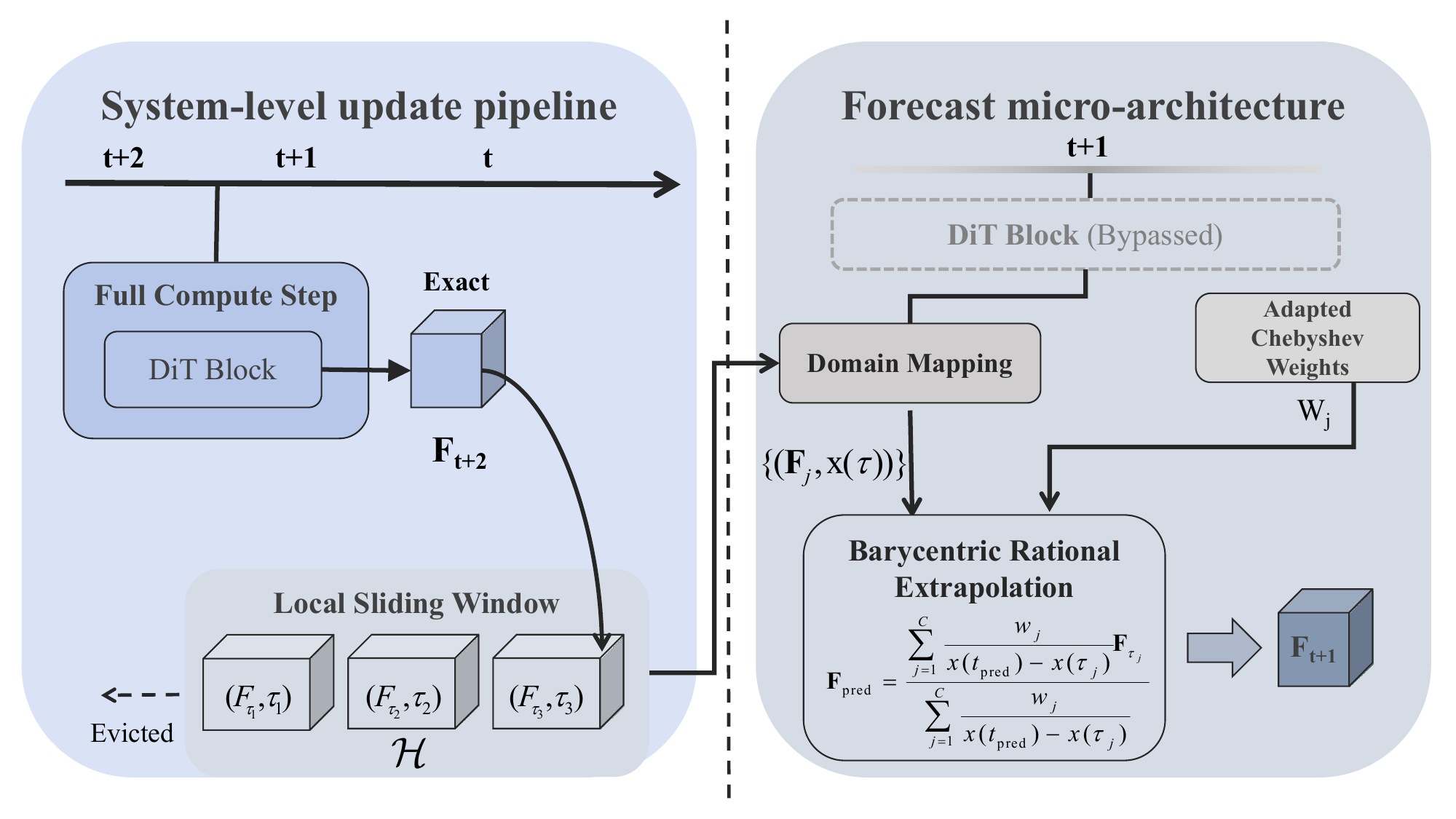}

\caption{Overview of the BRACE framework. 
Left: The system-level pipeline manages a FIFO Sliding Window to maintain a sparse historical cache $\mathcal{H}$ of exact DiT features. 
Right: The forecast micro-architecture bypasses neural network computation by synthesizing future features through Domain Mapping and Barycentric Rational Extrapolation, guided by Adapted Chebyshev Weights with negligible overhead while maintaining high fidelity.
}
\label{fig:brace_overview}
\Description{Two-part diagram of the BRACE pipeline. On the left, a FIFO sliding window stores recent DiT features from full compute steps. On the right, cached features are normalized and used in a barycentric rational extrapolation module to predict future features without running the network.}
\end{figure*}

Driven by the numerical insights from our empirical analysis, Barycentric Rational Forecasting with Chebyshev Enhancement (BRACE) is proposed, a training-free framework where the forecasting paradigm is shifted from derivative-driven polynomials to feature-driven rational aggregation. As illustrated in Figure~\ref{fig:brace_overview}, the inference process alternates at a fixed interval $k$ between full computation and rapid forecasting, a procedure formally detailed in Algorithm~\ref{alg:hbr_cache}.

During full computation steps, BRACE couples exact intermediate features $\mathbf{F}_\tau$ and their timesteps into state tuples $\mathbf{S} \triangleq (\mathbf{F}_\tau, \tau)$, pushing them into a FIFO Local Sliding Window. This cache $\mathcal{H}$ preserves the historical context on the feature manifold. 

Conversely, when a skip interval is triggered, BRACE activates its rational prediction mechanism. Specifically, a Domain Mapping operation first normalizes the cached timestamps into a canonical interval to ensure numerical stability. Subsequently, these mapped states are then synthesized into the predicted feature $\mathbf{F}_t$ via the Barycentric Rational Extrapolation module, which leverages Adapted Chebyshev Weights to robustly accommodate the potential non-smoothness identified in our earlier analysis.
\begin{algorithm}[t]
\caption{BRACE Sequential Inference (per layer $l$)}
\label{alg:hbr_cache}
\begin{algorithmic}[1]
\Require Total timesteps $T$, capacity $C$, interval $k$
\Statex \textbf{Internal State:} Local Sliding Window $\mathcal{H} \gets \emptyset$ 

\For{$t = T$ \textbf{down to} $1$} \Comment{Discrete reverse diffusion trajectory}
    \If {$t \pmod k == 0$} \Comment{Full Compute Step}
        \State $\mathbf{F}^l_t \gets \text{ForwardBlock}_l(\text{input})$ 
        \State $\mathcal{H} \leftarrow \text{Update}(\mathcal{H}, (\mathbf{F}^l_t, t), C)$ \Comment{FIFO strategy}
        \State $\tilde{\mathbf{F}}^l_t \gets \mathbf{F}^l_t$
    \Else \Comment{Forecast micro-architecture}
        \State $x(t) \gets \text{DomainMapping}(t, \mathcal{H})$ \Comment{Yields $|x(t)| > 1$}
        \State $\mathbf{F}^l_{pred} \gets \text{BarycentricRationalExtrap}(\mathcal{H}, x(t), W_j)$
        \State $\tilde{\mathbf{F}}^l_t \gets \mathbf{F}^l_{pred}$ \Comment{Bypass DiT Block}
    \EndIf
\EndFor
\end{algorithmic}
\end{algorithm}
\subsubsection{Local Sliding Window (FIFO Queue)} 
Serving as the foundation for state caching, BRACE employs a fixed-capacity FIFO queue to manage the local historical context. Formally, let $\mathbf{S}_j \triangleq (\mathbf{F}_{\tau_j}, \tau_j)$ denote a state tuple. At any given inference step, the active sliding window of capacity $C$ is defined as an ordered sequence:
\begin{equation}
    \mathcal{H} = (\mathbf{S}_1, \mathbf{S}_2, \dots, \mathbf{S}_C).
\end{equation}
Upon a full computation step that yields a new state $\mathbf{S}_{new}$, the FIFO mechanism updates the cache via a shift-and-append operation:
\begin{equation}
    \mathcal{H} \leftarrow (\mathbf{S}_2, \mathbf{S}_3, \dots, \mathbf{S}_C, \mathbf{S}_{new}),
\end{equation}
where the oldest state $\mathbf{S}_1$ is evicted. 

This finite-capacity design strategically confines the forecasting horizon to a sparse local window, effectively decoupling temporally decayed features while ensuring the trajectory satisfies local smoothness. As a result, by maintaining a strictly local cache, BRACE structurally bypasses complex global nonlinearities that could otherwise destabilize the rational predictor.

\subsubsection{Canonical Domain Mapping} 
To strictly align the diffusion timesteps with the theoretical foundations of the adapted weights, a Canonical Domain Mapping is prepended to the extrapolation process on the active cache $\mathcal{H}$. Specifically, it transforms each raw state tuple $\mathbf{S}_j = (\mathbf{F}_{\tau_j}, \tau_j)$ into a canonical state tuple $(\mathbf{F}_{\tau_j}, x(\tau_j))$ via an affine mapping. Mathematically, this transformation is essential because the numerical stability of the rational formulation is optimally preserved within the standard interval $[-1, 1]$. To project the raw timesteps into this standard domain, the mapping boundaries are determined dynamically by the active cache:
\begin{equation}
x(\tau) = 2 \cdot \frac{\tau - \tau_{\min}}{\tau_{\max} - \tau_{\min}} - 1,
\end{equation}
where $x(\tau_j) \in [-1, 1]$ for all cached states. By binding features directly to their canonical coordinates, this mapping ensures scale-invariance across different diffusion schedulers, while providing well-defined theoretical error bounds and preventing numerical instabilities during finite-precision inference.
\subsubsection{Barycentric Rational Extrapolation} 
Following the domain mapping stage, the mapped cache $\mathcal{H} = \{(\mathbf{F}_{\tau_j}, x(\tau_j))\}_{j=1}^{C}$ is aggregated to synthesize the future feature $\mathbf{F}_{\text{pred}}$ at the target timestep $t_{\text{pred}}$. Specifically, we employ the second barycentric form. By assigning fixed weights independent of node intervals, it is fundamentally transformed into a stable rational function.

\begin{equation}
    \mathbf{F}_{\text{pred}} = \frac{\sum_{j=1}^{C} \frac{w_j}{x(t_{\text{pred}}) - x(\tau_j)} \mathbf{F}_{\tau_j}}{\sum_{j=1}^{C} \frac{w_j}{x(t_{\text{pred}}) - x(\tau_j)}},
\label{eq:barycentric_formula}
\end{equation}
where $w_j$ denotes the adapted Chebyshev weights configured for numerical robustness and stable long-step extrapolation.

Leveraging this rational structure, the prediction is decomposed into a feature aggregation numerator and a self-normalizing denominator. This formulation provides two critical advantages: it inherently bounds predicted features to prevent polynomial divergence via self-normalization, and intrinsically absorbs localized non-smoothness by aggregating historical states rather than relying on unstable derivatives. Consequently, this design secures robust, high-fidelity long-step extrapolation.
\subsubsection{Adapted Chebyshev Weights}
The stability and approximation accuracy of Eq.~\eqref{eq:barycentric_formula} are determined by the rational weights $w_j$. In this context, it is well known that classical approximation theory shows that Chebyshev–Gauss–Lobatto nodes offer superior numerical stability and possess well-defined analytical weights~\cite{berrut2004barycentric}. Inspired by the numerical stability of classical Chebyshev weights, we formulate an adapted weighting scheme tailored to the nonlinear manifolds of DiT features:
\begin{equation}
    w_j = (-1)^{j+1} \cdot \delta_j, \quad \delta_j = 
    \begin{cases} 
        \frac{1}{2} & j = 1, \\ 
        \gamma & j = C, \\ 
        1 & \text{otherwise}. 
    \end{cases}
    \label{eq:adapted_weights}
\end{equation}

This formulation incorporates two critical structural designs. First, the alternating sign $(-1)^{j+1}$ serves as an inherent stabilizer to prevent numerical collapse in the extrapolation regime. Second, a boundary sensitivity coefficient $\gamma$ replaces the standard Lobatto terminal weight to flexibly accommodate varying model dynamics.

While applying these weights typically requires non-uniform Chebyshev sampling ($x_k = -\cos(\frac{k-1}{C-1}\pi)$), BRACE intentionally restricts the cache to an ultra-sparse capacity ($C \le 3$) to prevent long-range feature corruption. Crucially, in this low-order regime, these optimal nodes are mathematically identical to equidistant points. This equivalence enables theoretically optimal sampling via a simple fixed interval, thereby eliminating complex dynamic adjustments and circumventing the numerical instability that traditionally plagues barycentric extrapolation~\cite{webb2012stability}.
\subsection{Error Propagation and Stability Analysis}
To rigorously establish the numerical stability of BRACE under large skip intervals, we analyze its pointwise extrapolation error $\mathbf{E}(x) = \mathbf{F}(x) - \mathbf{F}{\text{extrap}}(x)$. Assume that the feature trajectory $\mathbf{F}$ is Lipschitz continuous on the inference domain, such that $|\mathbf{F}(x) - \mathbf{F}(x_i)| \le L |x - x_i|$, where $L$ is the Lipschitz constant (equivalent to the supremum of the first-order derivative $|\mathbf{F}'|{\infty}$). Substituting this Lipschitz condition into the barycentric error identity yields a robust upper bound on the error norm.
\begin{equation}
    \|\mathbf{E}(x)\| = \left\| \frac{\sum_{i=1}^C \frac{w_i}{x - x_i} (\mathbf{F}(x) - \mathbf{F}(x_i))}{\sum_{j=1}^C \frac{w_j}{x - x_j}} \right\| \le \frac{\sum_{i=1}^C |w_i| \cdot L}{|D(x)|}
    \label{eq:weighted_derivative_form}
\end{equation}
where $D(x) = \sum_{j=1}^C \frac{w_j}{x - x_j}$ denotes the barycentric denominator.

For extrapolation at target $x = x_C + s$ ($s > 0$), BRACE's uniformly spaced cached nodes yield $x_j = x_C - (C - j)k$ with a constant interval $k$, which inherently bounds the extrapolation range to $s \le k$. Substituting these reformulates $D(x)$ concerning $s$.
\begin{equation}
    D(s) = \sum_{i=1}^C \frac{w_i}{s + (C-i)k} = \frac{1}{k} \sum_{i=1}^C \frac{w_i}{\sigma + C - i}
\end{equation}
where $\sigma = s/k \in (0, 1]$ represents the relative skip ratio. Crucially, the summation term $\sum_{i=1}^C w_i / (\sigma + C - i)$ is strictly bounded below by a non-vanishing structural constant $\lambda > 0$ for all $\sigma \in (0, 1]$. A rigorous formal proof of this property is provided in the Supplemental Material.  Consequently, by incorporating the factor $1/k$ from the definition of $D(s)$, it holds that $|D(s)| \ge \lambda/k$, thereby directly establishing the localized error bound:
\begin{equation}
    \|\mathbf{E}(x_C + s)\| \le \frac{\sum_{i=1}^C |w_i| \cdot L}{\lambda} \cdot k
    \label{eq:generalized_error_bound_k}
\end{equation}
This suggests that the extrapolation error in BRACE is theoretically governed by the interval $k$ and the first-order derivative (Lipschitz constant $L$). This advantage is highly pronounced compared to conventional $m$-th order Taylor extrapolation (Eq.~\ref{eq:taylor}): for a prediction offset $s$, Taylor methods incur a truncation error of $\mathcal{O}(s^{m+1} \|\mathbf{F}^{(m+1)}\|_{\infty})$, leading to potential polynomial divergence and unbounded amplification of high-order derivatives during sharp transitions. By bypassing these numerical hazards and theoretically bounding the error via $k$ and $L$, BRACE ensures predictable stability even under aggressive skipping.

%% file: sec_4_Experiments.tex
\definecolor{highlightrow}{gray}{0.92}
\section{Experiments}

\subsection{Experimental Setup}
BRACE is evaluated across three representative generative tasks: (i) class-conditional image generation on ImageNet-1K~\cite{deng2009imagenet} using DiT-XL/2~\cite{peebles2023scalablediffusionmodelstransformers}; (ii) text-to-image generation on DrawBench~\cite{saharia2022photorealistictexttoimagediffusionmodels} via the rectified flow-based FLUX.1-dev~\cite{labs2025flux1kontextflowmatching,flux2024,liu2022flowstraightfastlearning}; and (iii) text-to-video generation with HunyuanVideo~\cite{kong2024hunyuanvideo} on the HunyuanLarge architecture. To provide a holistic assessment, all experiments are conducted under official inference protocols and default sampling parameters, with caching methods compared at identical acceleration ratios for fair visualization. Quantitative performance is assessed using FID-50K~\cite{fid}, sFID, and Inception Score for DiT-XL/2; ImageReward~\cite{xu2023imagereward}, CLIP Score~\cite{hessel2022clipscorereferencefreeevaluationmetric}, CycleReward~\cite{Bahng_2025_ICCV}, PSNR, SSIM~\cite{PSNRSSIM}, and LPIPS~\cite{LPIPsZhang_2018_CVPR} for FLUX.1-dev; and the 16-dimensional VBench suite~\cite{huang2023vbenchcomprehensivebenchmarksuite} for HunyuanVideo. Finally, computational efficiency is measured via relative FLOPs and latency against baselines.

\subsection{Class-Conditional Image Generation}

\label{sec:exp_discussion}
\begin{figure}[t]
    \centering
    \includegraphics[width=\linewidth]{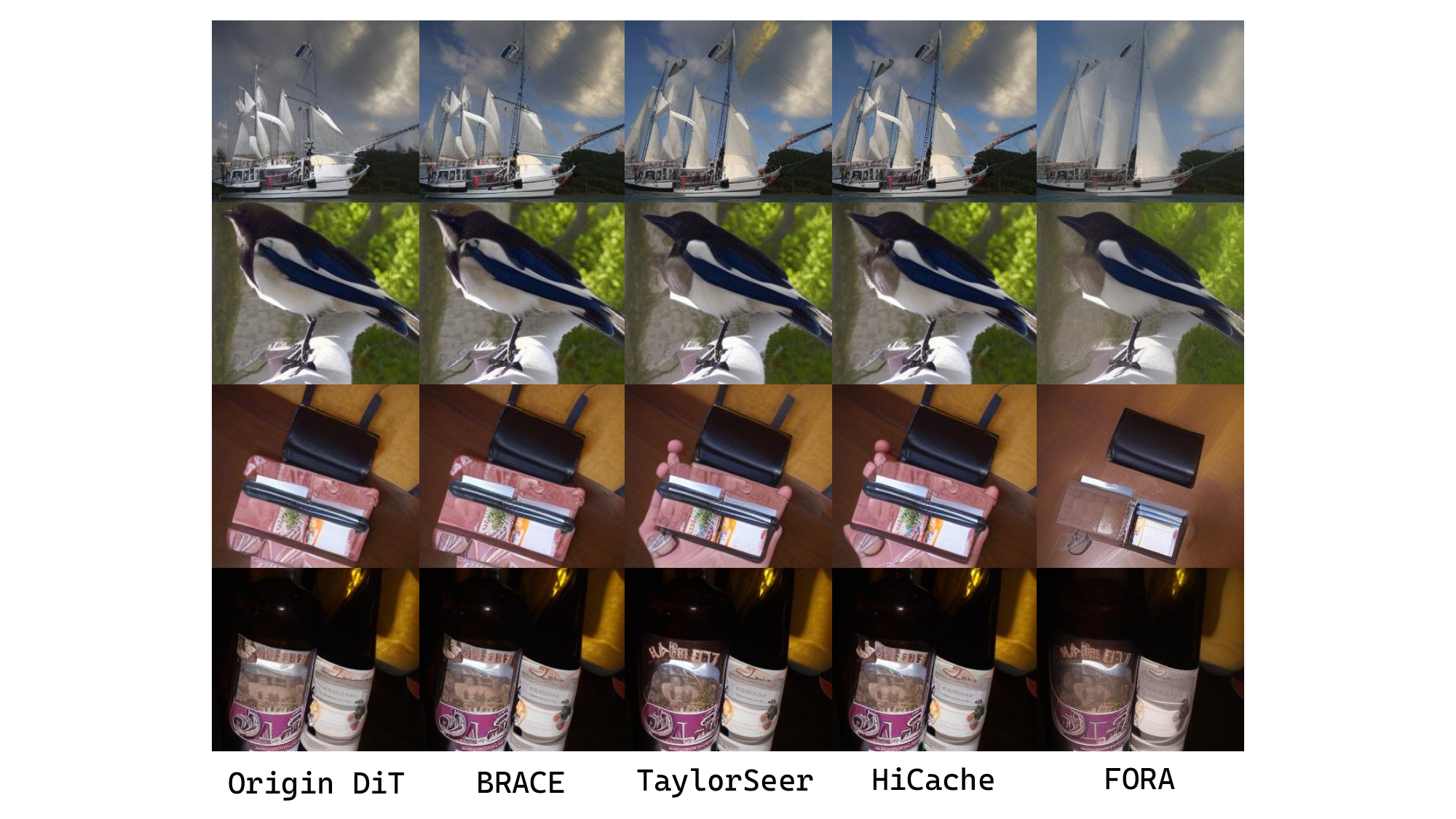}
    \caption{Qualitative comparison of different acceleration methods on DiT-XL/2}
    \label{fig:dit_viz_methods}
    \Description{A wide-span visual comparison showing that BRACE outperforms other methods in preserving geometric and semantic details on DiT.}
\end{figure}
\begin{table}[t]
    \centering
    \caption{Quantitative comparison of class-conditional image generation on ImageNet with DiT-XL/2}
    \label{tab:main_results}
    \footnotesize                  
    \setlength{\tabcolsep}{1.5pt}  
    \resizebox{\columnwidth}{!}{   
    \begin{tabular}{l c c c c c c}
        \toprule
        \textbf{Method}
        & \textbf{Lat. (s)$\downarrow$}
        & \textbf{FLOPs (T)$\downarrow$}
        & \textbf{Spd.$\uparrow$}
        & \textbf{FID$\downarrow$}
        & \textbf{sFID$\downarrow$}
        & \textbf{IS$\uparrow$} \\
        \midrule
        DDIM-50 steps~\cite{song2022denoisingdiffusionimplicitmodels} & 1.18 & 23.74 & 1.00$\times$ & 2.29 & 4.29 & 237.31 \\
        DDIM-25 steps~\cite{song2022denoisingdiffusionimplicitmodels} & 0.62 & 11.87 & 2.00$\times$ & 2.91 & 4.57 & 232.22 \\
        \midrule
        DDIM-15 steps~\cite{song2022denoisingdiffusionimplicitmodels} & 0.42 & 7.12 & 3.33$\times$ & 5.09 & 6.11 & 205.54 \\
        FORA($N$=4)~\cite{selvaraju2024FORAfastforwardcachingdiffusion} & 0.45 & 6.66 & 3.56$\times$ & 4.73 & 8.45 & 214.20 \\
        Taylor ($N$=4, $O$=4)~\cite{liu2025reusingforecastingacceleratingdiffusion} & 0.61 & 6.66 & 3.56$\times$ & 2.59 & 5.18 & 232.90 \\
        HiCache ($N$=4, $O$=4)~\cite{feng2026hicachepluginscaledhermiteupgrade} & 0.57 & 6.66 & 3.56$\times$ & 2.51 & 5.20 & 232.72 \\
        \rowcolor{highlightrow}
        \textbf{BRACE($N$=4,$C$=3,$\gamma$=0.5)} & 0.46 & \textbf{6.66} & \textbf{3.56}$\times$ & \textbf{2.46} & \textbf{4.90} & \textbf{233.49} \\
        \midrule
        DDIM-12 steps~\cite{song2022denoisingdiffusionimplicitmodels} & 0.33 & 5.70 & 4.17$\times$ & 7.94 & 8.07 & 183.88 \\
        FORA($N$=5)~\cite{selvaraju2024FORAfastforwardcachingdiffusion} & 0.37 & 5.24 & 4.53$\times$ & 5.81 & 9.61 & 199.96 \\
        Taylor ($N$=5, $O$=4)~\cite{liu2025reusingforecastingacceleratingdiffusion} & 0.55 & 5.24 & 4.53$\times$ & 2.73 & 5.34 & 228.75 \\
        HiCache ($N$=5, $O$=4)~\cite{feng2026hicachepluginscaledhermiteupgrade} & 0.50 & 5.24 & 4.53$\times$ & 2.67 & 5.48 & 229.44 \\
        \rowcolor{highlightrow}
        \textbf{BRACE($N$=5,$C$=3,$\gamma$=0.5)} & 0.42 & \textbf{5.24} & \textbf{4.53}$\times$ & \textbf{2.59} & \textbf{5.03} & \textbf{228.97} \\
        \midrule
        DDIM-10 steps~\cite{song2022denoisingdiffusionimplicitmodels} & 0.30 & 4.75 & 5.00$\times$ & 12.17 & 11.24 & 159.20 \\
        FORA($N$=6)~\cite{selvaraju2024FORAfastforwardcachingdiffusion} & 0.34 & 4.76 & 4.98$\times$ & 9.22 & 14.86 & 169.02 \\
        Taylor ($N$=6, $O$=4)~\cite{liu2025reusingforecastingacceleratingdiffusion} & 0.50 & 4.76 & 4.98$\times$ & 3.17 & 6.36 & \textbf{222.00} \\
        HiCache ($N$=6, $O$=4)~\cite{feng2026hicachepluginscaledhermiteupgrade} & 0.46 & 4.76 & 4.98$\times$ & 3.09 & 6.30 & \textbf{220.55} \\
        \rowcolor{highlightrow}
        \textbf{BRACE($N$=6,$C$=3,$\gamma$=0.5)} & 0.39 & \textbf{4.76} & \textbf{4.98}$\times$ & \textbf{3.06} & \textbf{5.75} & 218.20 \\
        \bottomrule
    \end{tabular}
    }
\end{table}

\paragraph{Quantitative Analysis}
As demonstrated in Table~\ref{tab:main_results}, BRACE consistently outperforms all baselines in both generative fidelity and efficiency across various skip intervals, achieving the lowest FID and sFID in every configuration. This advantage stems from our second-form barycentric framework, which inherently tames the trajectory irregularities that plague conventional methods. Specifically, while derivative-driven polynomial approaches~\cite{liu2025reusingforecastingacceleratingdiffusion,feng2026hicachepluginscaledhermiteupgrade} suffer from extrapolation instability—often leading to structural vulnerabilities and ghosting artifacts—BRACE maintains superior geometric integrity. By bypassing volatile high-order derivative estimations, BRACE strikes a favorable balance between mathematical expressiveness and numerical robustness, providing a more stable alternative for feature extrapolation in accelerated diffusion inference.

\paragraph{Qualitative Comparison}
Figure~\ref{fig:dit_viz_methods} compares different methods on DiT-XL/2 under matched acceleration settings. FORA~\cite{selvaraju2024FORAfastforwardcachingdiffusion} produces over-smoothing and color shifts, whereas TaylorSeer and HiCache~\cite{liu2025reusingforecastingacceleratingdiffusion,feng2026hicachepluginscaledhermiteupgrade} exhibit texture and structural distortions, particularly in Row~3. In contrast, BRACE better preserves clean backgrounds, sharp details, and semantic structure.

\subsection{Text-to-Image Generation}
\begin{figure*}[t]
    \centering
    \includegraphics[width=\textwidth]{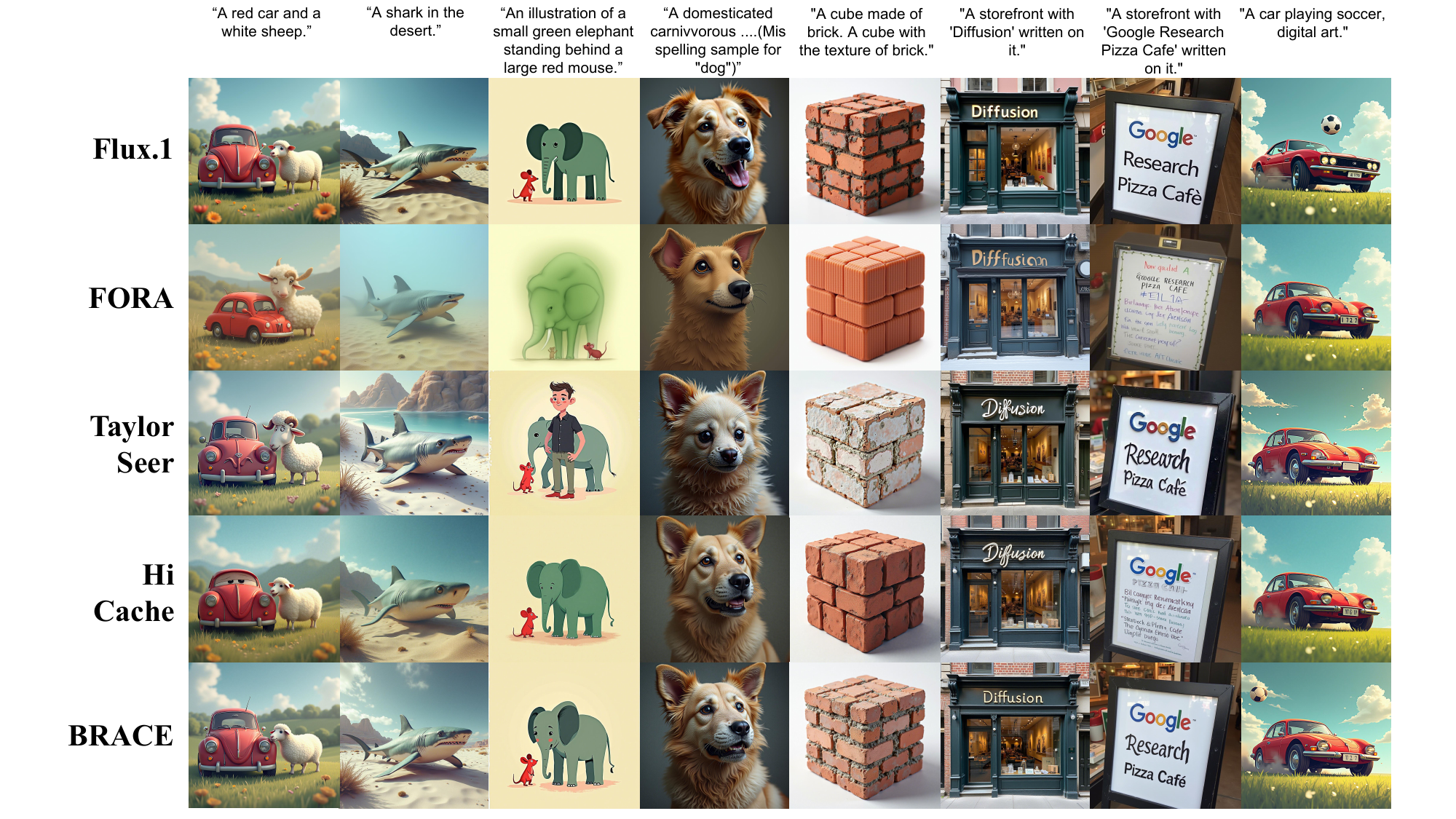}
    \caption{Visualization examples for different acceleration methods on Flux.}
    \label{fig:flux_viz_methods}
    \Description{Visual comparison of image generation results produced by various acceleration methods on the Flux model.}
\end{figure*}
\paragraph{Quantitative Analysis}
Table~\ref{tab:main_flux_updated} reports the results on FLUX.1-dev~\cite{flux2024}, where BRACE achieves a new state-of-the-art in both semantic alignment and structural fidelity. Across all skip intervals, BRACE consistently yields the highest ImageReward~\cite{xu2023imagereward}, CLIP Score~\cite{hessel2022clipscorereferencefreeevaluationmetric}, CycleReward~\cite{Bahng_2025_ICCV}; Furthermore, BRACE achieves the lowest LPIPS and highest SSIM in nearly all settings, effectively confirming that our rational extrapolation effectively preserves intricate fine-grained geometric structures and minimizes perceptual distortion compared to derivative-driven baselines.

\paragraph{Qualitative Comparison}
Figure~\ref{fig:flux_viz_methods} compares BRACE with baseline methods on FLUX.1-dev~\cite{flux2024} at an extreme 5.55$\times$ speedup. Under this challenging setting, TaylorSeer~\cite{liu2025reusingforecastingacceleratingdiffusion} exhibits semantic and text-structure distortions, HiCache~\cite{feng2026hicachepluginscaledhermiteupgrade} causes over-smoothing and blurred text, and FORA~\cite{selvaraju2024FORAfastforwardcachingdiffusion} produces noticeable typographical errors. In column~6, the baselines corrupt the double ``f'' by merging or deforming the characters, whereas BRACE preserves sharp, legible typography consistent with the full-step baseline.
\begin{table*}[t]
  \centering
  \caption{Quantitative comparison of text-to-image generation on FLUX}
  \label{tab:main_flux_updated}
  \setlength{\tabcolsep}{2.6pt} 
  \small 
  \begin{tabular}{lccccccccc}
    \toprule
    \textbf{Method}
    & \textbf{Latency(s)$\downarrow$}
    & \textbf{FLOPs(T)$\downarrow$}
    & \textbf{Speed$\uparrow$}
    & \textbf{Img. Reward$\uparrow$}
    & \textbf{CLIP Score$\uparrow$}
    & \textbf{Cycle Reward$\uparrow$}
    & \textbf{PSNR$\uparrow$}
    & \textbf{SSIM$\uparrow$}
    & \textbf{LPIPS$\downarrow$} \\
    \midrule
    FLUX.1 [dev] - 50 steps$^\dagger$
    & 17.18 & 3719.50 & 1.00$\times$ & 0.9898 & 27.4611 & 0.8800 & $\infty$ & 1.0000 & 0.0000 \\
    FLUX.1 [dev] - 25 steps
    & 8.80 & 1859.75 & 2.00$\times$ & 0.9375 & 27.3388 & 0.8466 & 19.2599 & 0.7749 & 0.3583 \\
    FLUX.1 [dev] - 20 steps
    & 7.12 & 1487.80 & 2.62$\times$ & 0.9387 & 27.2505 & 0.8466 & 18.1371 & 0.7484 & 0.4006 \\
    \midrule
    FORA ($\mathcal{N}$=5)~\cite{selvaraju2024FORAfastforwardcachingdiffusion}
    & 5.01 & 893.54 & 4.16$\times$ & 0.8253 & 27.2377 & 0.8353 & 15.9899 & 0.6677 & 0.5206 \\
    TaylorSeer ($\mathcal{N}$=5, $\mathcal{O}$=1)~\cite{liu2025reusingforecastingacceleratingdiffusion}
    & 5.14 & 893.54 & 4.16$\times$ & 0.9919 & 27.5143 & 0.8540 & 18.4497 & 0.7424 & 0.4146 \\
    HiCache ($\mathcal{N}$=5, $\mathcal{O}$=1)~\cite{feng2026hicachepluginscaledhermiteupgrade}
    & 5.12 & 893.54 & 4.16$\times$ & 0.9350 & 27.4034 & 0.8346 & 19.2595 & 0.7503 & 0.3911 \\
    \rowcolor{gray!10}
    \textbf{BRACE$^\ddagger$ ($\mathcal{N}$=5, $\mathcal{C}$=2, $\gamma$=0.7)}
    & 5.04 & 893.54 & 4.16$\times$ & \textbf{1.0021} & \textbf{27.5886} & \textbf{0.8596} & \textbf{19.3671} & \textbf{0.7598} & \textbf{0.3770} \\
    \midrule
    FORA ($\mathcal{N}$=6)~\cite{selvaraju2024FORAfastforwardcachingdiffusion}
    & 4.40 & 744.81 & 4.99$\times$ & 0.7836 & 27.1541 & 0.8087 & 15.7298 & 0.6663 & 0.5253 \\
    TaylorSeer ($\mathcal{N}$=6, $\mathcal{O}$=1)~\cite{liu2025reusingforecastingacceleratingdiffusion}
    & 4.57 & 744.81 & 4.99$\times$ & 0.9850 & 27.5199 & 0.8480 & 17.5396 & 0.7078 & 0.4657 \\
    HiCache ($\mathcal{N}$=6, $\mathcal{O}$=1)~\cite{feng2026hicachepluginscaledhermiteupgrade}
    & 4.55 & 744.81 & 4.99$\times$ & 0.9427 & 27.6898 & 0.8539 & 18.5159 & 0.7155 & 0.4503 \\
    HiCache ($\mathcal{N}$=6, $\mathcal{O}$=2)~\cite{feng2026hicachepluginscaledhermiteupgrade}
    & 5.03 & 744.81 & 4.99$\times$ & 0.9892 & 27.6586 & 0.8428 & 18.5044 & 0.7268 & 0.4316 \\
    \rowcolor{gray!10}
    \textbf{BRACE$^\ddagger$ ($\mathcal{N}$=6, $\mathcal{C}$=2, $\gamma$=0.7)}
    & 4.43 & 744.81 & 4.99$\times$ & \textbf{1.0025} & \textbf{27.7246} & \textbf{0.8709} & \textbf{18.6275} & \textbf{0.7285} & \textbf{0.4246} \\
    \midrule
    FORA ($\mathcal{N}$=7)~\cite{selvaraju2024FORAfastforwardcachingdiffusion}
    & 4.10 & 670.44 & 5.55$\times$ & 0.7289 & 27.0135 & 0.7962 & 15.5290 & 0.6548 & 0.5437 \\
    TaylorSeer ($\mathcal{N}$=7, $\mathcal{O}$=1)~\cite{liu2025reusingforecastingacceleratingdiffusion}
    & 4.31 & 670.44 & 5.55$\times$ & 0.9420 & 27.4039 & 0.7970 & 16.8563 & 0.6844 & 0.5062 \\
    HiCache ($\mathcal{N}$=7, $\mathcal{O}$=1)~\cite{feng2026hicachepluginscaledhermiteupgrade}
    & 4.20 & 670.44 & 5.55$\times$ & 0.9191 & 27.6949 & 0.8494 & \textbf{18.2778} & 0.7032 & 0.4708 \\
    HiCache ($\mathcal{N}$=7, $\mathcal{O}$=2)~\cite{feng2026hicachepluginscaledhermiteupgrade}
    & 4.72 & 670.44 & 5.55$\times$ & 0.9881 & 27.5651 & 0.8384 & 17.9911 & 0.7098 &  0.4592 \\
    \rowcolor{gray!10}
    \textbf{BRACE$^\ddagger$ ($\mathcal{N}$=7, $\mathcal{C}$=2, $\gamma$=0.7)}
    & 4.12 & 670.44 & 5.55$\times$ & \textbf{0.9884} & \textbf{27.7372} & \textbf{0.8626} & 18.1925 & \textbf{0.7137} & \textbf{0.4491} \\
    \bottomrule
  \end{tabular}
\end{table*}
\subsection{Text-to-Video Generation}
\begin{table}[t]
  \centering
  \caption{Quantitative comparison of text-to-video generation on HunyuanVideo}
  \label{tab:t2v_generation}
  \footnotesize                  
  \setlength{\tabcolsep}{1.5pt}  
  \resizebox{\columnwidth}{!}{   
  \begin{tabular}{lccccc}
    \toprule
    \textbf{Method} & \textbf{Lat. (s)$\downarrow$} & \textbf{Spd.$\uparrow$} & \textbf{FLOPs (T)$\downarrow$} & \textbf{Spd.$\uparrow$} & \textbf{VBench (\%)$\uparrow$} \\
    \midrule
    Original (50 steps) & 190.2 & 1.00$\times$ & 29773 & 1.00$\times$ & 80.68 \\
    DDIM (22\% steps)~\cite{song2022denoisingdiffusionimplicitmodels} & 42.7 & 4.55$\times$ & 6550 & 4.55$\times$ & 78.73 \\
    \midrule
    FORA (N=6)~\cite{selvaraju2024FORAfastforwardcachingdiffusion} & 37.1 & 5.13$\times$ & 5359 & 5.56$\times$ & 78.86 \\
    TaylorSeer (N=6, O=1)~\cite{liu2025reusingforecastingacceleratingdiffusion} & 37.8 & 5.03$\times$ & 5359 & 5.56$\times$ & 79.87 \\
    HiCache (N=6, O=1)~\cite{feng2026hicachepluginscaledhermiteupgrade} & 37.8 & 5.03$\times$ & 5359 & 5.56$\times$ & 79.93 \\
    \textbf{BRACE (N=6, C=2,$\gamma$=0.4)} & 37.9 & 5.03$\times$ & 5359 & 5.56$\times$ & \textbf{80.14} \\
    \bottomrule
  \end{tabular}
  }
\end{table}
\begin{figure*}[t]
    \centering
    \includegraphics[width=\textwidth]{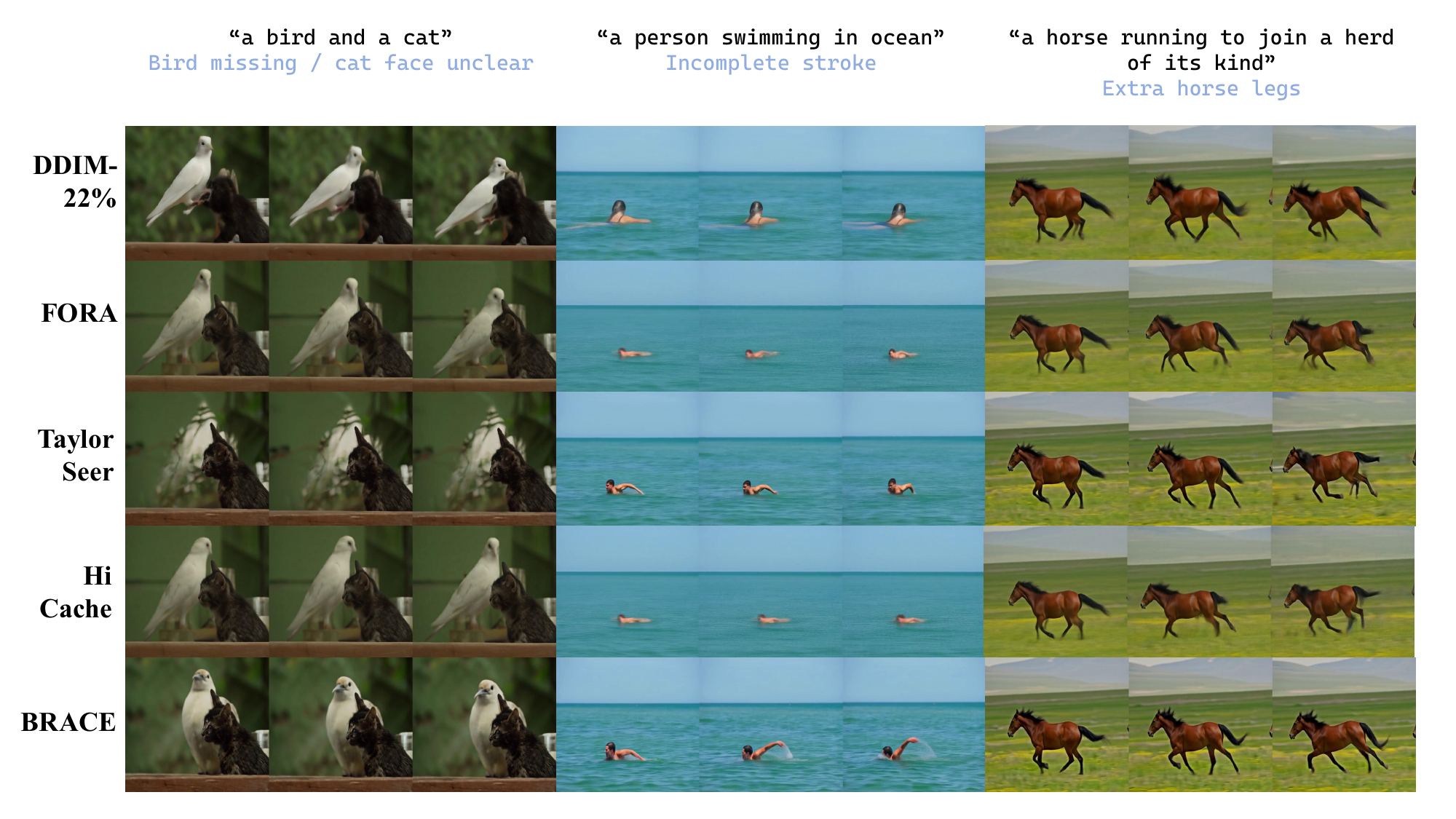}
    \caption{\textbf{Qualitative comparison on Text-to-Video Generation.}}
    \label{fig:video_comparison}
\end{figure*}
\paragraph{Quantitative Analysis}
Table~\ref{tab:t2v_generation} extends our evaluation to text-to-video generation, where BRACE establishes a new state-of-the-art in preserving spatiotemporal fidelity under extreme acceleration. Specifically, at a high skip interval ($N=6$), BRACE achieves the highest VBench score of 80.14, significantly outperforming both direct reuse (FORA~\cite{selvaraju2024FORAfastforwardcachingdiffusion}: 78.8) and derivative-driven forecasting methods like TaylorSeer~\cite{liu2025reusingforecastingacceleratingdiffusion} (79.87) and HiCache~\cite{feng2026hicachepluginscaledhermiteupgrade} (79.93). Notably, it successfully recovers the performance level closest to the unaccelerated 50-step baseline, effectively minimizing the quality gap typically observed in high-speed video inference.
\paragraph{Qualitative Comparison}
Figure~\ref{fig:video_comparison} shows that TaylorSeer~\cite{liu2025reusingforecastingacceleratingdiffusion} and FORA~\cite{selvaraju2024FORAfastforwardcachingdiffusion} lose entities, facial details, or coherent motion in the bird--cat and swimming scenes. In the running-horse scene, TaylorSeer and HiCache~\cite{feng2026hicachepluginscaledhermiteupgrade} produce structural artifacts such as extra legs. Across these cases, the competing methods struggle to preserve both frame-level details and cross-frame dynamics. In contrast, BRACE better preserves semantic content, anatomical structure, and temporal consistency.
\subsection{Ablation Study}
\label{sec:ablation}
\begin{table}[t]
  \centering
  \caption{Performance on DiT-XL/2 using different window capacity}
  \label{tab:ablation_capacity_dit}
  \setlength{\tabcolsep}{6pt}
  \small
  \begin{tabular}{lcccc}
    \toprule
    \textbf{Method} & \textbf{Speed$\uparrow$} & \textbf{FID$\downarrow$} & \textbf{sFID$\downarrow$} & \textbf{IS$\uparrow$} \\
    \midrule
    DDIM-50~\cite{song2022denoisingdiffusionimplicitmodels} & 1.00$\times$ & 2.29 & 4.29 & 237.31 \\
    \midrule
    BRACE ($C=2$) & 3.56$\times$ & 2.53 & 5.15 & 231.45 \\
    \rowcolor{gray!10}
    \textbf{BRACE ($C=3$)} & \textbf{3.56$\times$} & \textbf{2.46} & \textbf{4.90} & \textbf{233.49} \\
    BRACE ($C=4$) & 3.56$\times$ & 2.58 & 5.32 &  234.19 \\
    \midrule
    HiCache ($N=4$)~\cite{feng2026hicachepluginscaledhermiteupgrade} & 3.56$\times$ & 2.51 & 5.20 & 232.72 \\
    TaylorSeer ($N=4$)~\cite{liu2025reusingforecastingacceleratingdiffusion}  & 3.56$\times$ & 2.59 & 5.18 & 232.90 \\
    \bottomrule
  \end{tabular}
\end{table}
\begin{table}[t]
  \centering
  \caption{Performance on FLUX using different boundary sensitivity}
  \label{tab:ablation_flux_comprehensive}
  \setlength{\tabcolsep}{2.5pt}
  \small
  \begin{tabular}{lccccc}
    \toprule
    \textbf{Method} & \textbf{Speed$\uparrow$} & \textbf{Reward$\uparrow$} & \textbf{PSNR$\uparrow$} & \textbf{SSIM$\uparrow$} & \textbf{LPIPS$\downarrow$} \\
    \midrule
   
    BRACE ($\gamma=0.4$)     & 4.16$\times$ & 0.9500 & 16.5315 & 0.6850 & 0.5150 \\
    BRACE ($\gamma=0.5$)     & 4.16$\times$ & 1.0004 & 18.4343 & 0.7421 & 0.4132 \\
    BRACE ($\gamma=0.6$)     & 4.16$\times$ & 0.9991 & 19.1061 & 0.7565 & 0.3840 \\
    BRACE ($\gamma=0.65$)    & 4.16$\times$ & 0.9972 & 19.2665 & 0.7587 & 0.3791 \\
     \rowcolor{gray!10}
    BRACE ($\gamma=0.7$)     & 4.16$\times$ & \textbf{1.0021} & 19.3671 & \textbf{0.7598} & 0.3770 \\
    BRACE ($\gamma=1.0$)     & 4.16$\times$ & 0.9995 & \textbf{19.5934} & \textbf{0.7598} & \textbf{0.3744} \\
    BRACE (Uniform)           & 4.16$\times$ & 0.8859 & 19.0233 & 0.7325 &  0.4236 \\
    BRACE (Berrut)           & 4.16$\times$ & 0.9938 & 18.4343 &  0.7421 & 0.4131 \\
    BRACE (Floater-Hormann)  & 4.16$\times$ & 0.9259 & 19.0242 & 0.7322 & 0.4133 \\  
    BRACE (First Form)       & 4.16$\times$ & 0.9830 & 18.4116  & 0.7408 &  0.4245 \\
    \midrule
    HiCache ($N$=5)~\cite{feng2026hicachepluginscaledhermiteupgrade}          & 4.16$\times$ & 0.9350 & 19.2595 & 0.7503 & 0.3911 \\
    TaylorSeer ($N$=5)~\cite{liu2025reusingforecastingacceleratingdiffusion}        & 4.16$\times$ & 0.9919 & 18.4497 & 0.7424 & 0.4146 \\
    FORA ($N$=5)~\cite{selvaraju2024FORAfastforwardcachingdiffusion}             & 4.16$\times$ & 0.8253 & 15.9899 & 0.6677 & 0.5206 \\
    \bottomrule
  \end{tabular}
\end{table}
To evaluate the impact of window capacity and boundary sensitivity on the performance of BRACE, we conduct comprehensive ablations on FLUX.1-dev and DiT-XL/2. Regarding window capacity (Table~\ref{tab:ablation_capacity_dit}), $C=3$ provides the optimal approximation of feature curvature. Further increasing $C$ to 4 leads to performance degradation primarily attributed to the accumulation of long-range extrapolation noise, while reducing $C$ to 2 also weakens the feature fitting capability and overall model performance. 

We further analyze the boundary sensitivity $\gamma$ (Table~\ref{tab:ablation_flux_comprehensive}). Reconstruction metrics (PSNR, SSIM) improve monotonically as $\gamma \to 1.0$. However, perceptual quality (\textit{ImageReward}) peaks at $\gamma=0.7$ and subsequently declines; thus, $\gamma=0.7$ is established as the default. A more systematic exploration of the underlying causal factors driving this divergence between reconstruction and perceptual metrics is therefore deferred to future work.

Beyond parameter tuning, we investigate different extrapolation schemes. As shown in Table~\ref{tab:ablation_flux_comprehensive}, while the \textit{First Barycentric Form} provides a reasonable baseline, it becomes susceptible to numerical drift under aggressive extrapolation. In contrast, our \textit{Second Barycentric Form} ensures robust structural stability. Within this rational framework, we benchmark our adapted Chebyshev weights against three classical schemes: Uniform, Berrut, and Floater-Hormann~\cite{floater2007barycentric}. While these provide only basic stability, our adapted weights ultimately achieve superior alignment with the nonlinear evolution of DiT features by effectively capturing the asymmetric temporal dynamics of the generative trajectory.

%% file: sec_5_conclusion.tex
\section{Conclusion}
We presented \textbf{BRACE}, a training-free acceleration method for diffusion transformers that revisits feature caching through the lens of sequence-based forecasting. Motivated by the inherent structural stability of barycentric rational interpolation, we abandon the reliance on conventional finite-difference derivative estimation and polynomial extrapolation---which frequently exhibit numerical instability around sharp irregularities. Instead, BRACE directly extrapolates cached hidden states using this barycentric rational form, ensuring predictable stability even under long skip intervals. Across DiT~\cite{peebles2023scalablediffusionmodelstransformers} class-conditional generation on ImageNet, FLUX.1~\cite{flux2024} text-to-image generation, and large-scale text-to-video benchmarks with HunyuanVideo~\cite{kong2024hunyuanvideo}, BRACE consistently achieves a better quality--efficiency trade-off than reuse-based caching and derivative-driven polynomial forecasting. By flexibly adapting to the diverse feature dynamics of different DiT architectures, it delivers substantial speedups while maintaining strong fidelity. We hope this work serves to expand the feature forecasting paradigm from a distinct perspective.